\documentclass{article}

\usepackage{microtype}
\usepackage{graphicx}
\usepackage{latexsym}
\usepackage{tikz}
\usepackage{rotating,caption,tabularx,booktabs}
\usetikzlibrary{bayesnet}
\usetikzlibrary{arrows}
\usepackage{color}
\usepackage{float}
\usepackage{bm}
\usepackage{multirow}
\usepackage{makecell}
\usepackage{algorithmic}
\usepackage{tabularx}
\usepackage[numbers]{natbib}
\usepackage{amsmath,amssymb}
\usepackage{subfigure}
\usepackage{booktabs} 
\def\approxprop{%
  \def\p{%
    \setbox0=\vbox{\hbox{$\propto$}}%
    \ht0=0.6ex \box0 }%
  \def\s{%
    \vbox{\hbox{$\sim$}}%
  }%
  \mathrel{\raisebox{0.7ex}{%
      \mbox{$\underset{\s}{\p}$}%
    }}%
}

\usepackage{hyperref}

\usepackage[accepted]{icml2020}

\icmltitlerunning{On the Variational Posterior of Dirichlet Process Deep Latent Gaussian Mixture Models}

\begin{document}

\twocolumn[
\icmltitle{On the Variational Posterior of Dirichlet Process  \\ Deep Latent Gaussian Mixture Models}

\icmlsetsymbol{equal}{*}

\begin{icmlauthorlist}
\icmlauthor{Amine Echraibi}{to,IMT}
\icmlauthor{Joachim Flocon-Cholet}{to}
\icmlauthor{St\'ephane Gosselin}{to}
\icmlauthor{Sandrine Vaton}{IMT}
\end{icmlauthorlist}

\icmlaffiliation{to}{Orange Labs, Lannion, France}
\icmlaffiliation{IMT}{Institute Mines-Telecom Atlantique, Brest, France}

\icmlcorrespondingauthor{Amine Echraibi}{amine.echraibi@orange.com}
\icmlcorrespondingauthor{Sandrine Vaton}{sandrine.vaton@imt-atlantique.fr}
\icmlcorrespondingauthor{Joachim Flocon-Cholet}{joachim.floconcholet@orange.com}
\icmlcorrespondingauthor{St\'ephane Gosselin}{stephane.gosselin@orange.com}

\icmlkeywords{Machine Learning, ICML}

\vskip 0.3in
]



\printAffiliationsAndNotice{}  

\begin{abstract}
Thanks to the reparameterization trick, deep latent Gaussian models have shown tremendous success recently in learning latent representations. The ability to couple them however with nonparametric priors such as the Dirichlet Process (DP) hasn't seen similar success due to its non parameterizable nature. In this paper, we present an alternative treatment of the variational posterior of the Dirichlet Process Deep Latent Gaussian Mixture Model (DP-DLGMM), where we show that the prior cluster parameters and the variational posteriors of the beta distributions and cluster hidden variables can be updated in closed-form. This leads to a standard reparameterization trick on the Gaussian latent variables knowing the cluster assignments. We demonstrate our approach on standard benchmark datasets, we show that our model is capable of generating realistic samples for each cluster obtained, and manifests competitive performance in a semi-supervised setting.
\end{abstract}

\section{Introduction}

Nonparametric Bayesian priors, such as the Dirichlet Process (DP), have been widely adopted in the probabilistic graphical community. Their ability to generate an infinite amount of probability distributions using a discrete latent variable makes them ideally suited for automatic model selection. The most famous applications of the DP have been however limited to classical probabilistic graphical models such as Dirichlet Process Mixture Models and Hierarchical Dirichlet Process Hidden Markov Models  \cite{blei2006variational, fox2008hdp, zhang2016stochastic}.

Recently, deep generative models such as Deep Latent Gaussian Models (DLGMs) and Variational AutoEncoders (VAEs) \cite{kingma2013auto,rezende2014stochastic} have shown huge success in modeling and generating complex data structures such as images. Various proposals to generalize these models to the mixture and nonparametric mixture cases have been made \cite{nalisnick2016approximate,nalisnick2016stick,dilokthanakul2016deep,jiang2016variational}. Introducing such priors on top of the deep generative model can improve its generative capabilities, preserve class structure in the latent representation space, and offer a nonparametric way of performing model selection with respect to the size of the generative model.

The main challenge posed by such models lies in the inference process. Deep generative models with continuous latent variables owe their success mainly to the reparameterization trick \cite{kingma2013auto,rezende2014stochastic}. This approach provides an efficient and scalable method for obtaining low variance estimates of the gradient of the variational lower bound with respect to variational posterior parameters. Applying this approach directly to the variational posterior of the DP is not straightforward, due to the fact that a reparameterization trick for the beta distributions is hard to obtain \cite{ruiz2016generalized}. One approach to bypass this issue have been proposed by  \cite{nalisnick2016stick}, where the authors used the Kumaraswamy distribution \cite{kumaraswamy1980generalized} as a higher entropy alternative for the beta distribution in the variational posterior. However, by deriving the nature of the variational posterior directly from the variational lower bound, we can show that the appropriate distribution is in fact the beta distribution.

In this paper we provide an alternative treatment of the variational posterior of the DP-DLGMM, where we combine classical variational inference to derive the variational posteriors of the beta distributions and cluster hidden variables, and neural variational inference for the hidden variables of the latent Gaussian model. This leads to gradient ascent updates over the parameters present in nonlinear transformations where the reparameterization trick can be applied knowing the cluster assignment. As for the remaining parameters, closed-form solutions can be obtained by maximization of the evidence lower bound.

\section{Dirichlet Process Deep Latent Gaussian Mixture Models}\label{DPDLGMM}
\begin{figure}[t]
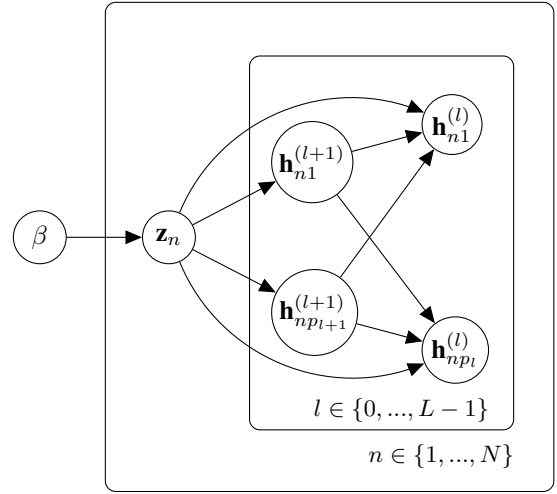

  \tikz{
    
    \node[latent](beta){$\beta$};
     
    \node[latent, right= of beta] (zn) {$\textbf{z}_n$};
    \node[latent, right=of zn, yshift=+1cm](h_n_l1) {$\textbf{h}_{n1}^{(l+1)}$};
    \node[latent, right=of zn, yshift=-1cm](h_n_lp) {$\textbf{h}_{np_{l+1}}^{(l+1)}$};
    \node[latent, right=of zn, yshift=+1.5cm, xshift=2cm](h_n_t1) {$\textbf{h}_{n1}^{(l)}$};
    \node[latent, right=of zn, yshift=-1.5cm, xshift=2cm](h_n_tp) {$\textbf{h}_{np_{l}}^{(l)}$};
     \edge {beta}{zn};
     \edge{zn}{h_n_l1};
     \edge{zn}{h_n_lp};
     \edge{h_n_l1}{h_n_t1};
     \edge{h_n_l1}{h_n_tp};
     \edge{h_n_lp}{h_n_t1};
     \edge{h_n_lp}{h_n_tp};
     
     \path[every node/.style={font=\sffamily\small}]
     (zn) edge[bend left=45, ->] node [left] {} (h_n_t1);
     \path[every node/.style={font=\sffamily\small}]
     (zn) edge[bend right=45, ->] node [left] {} (h_n_tp);
     
      \plate [inner sep=.3cm,xshift=.02cm,yshift=.2cm] {plate1} {(h_n_l1)(h_n_lp)(h_n_t1)(h_n_tp)} {$l \in \{0,...,L-1 \} $};
      \plate [inner sep=0.5cm,xshift=.02cm,yshift=.2cm] {plate2} {(plate1)(zn)} {$n \in \{1,...,N \}$};
     
     }
     \caption{The graphical representation of the generative process of the model, with the convention $\textbf{x} = \textbf{h}^{(0)}$.}
     \label{fig:plate}
\end{figure}

Generalizing deep latent Gaussian models to the Dirichlet process mixture case can be obtained by adding a Dirichlet process prior on the hidden cluster assignments. We denote these cluster assignments by $\mathbf{z}$. Following the assignment of a cluster hidden variable, a deep latent Gaussian model is defined for the assigned cluster similar to \cite{rezende2014stochastic}. We adopt the stick-breaking construction of the Dirichlet Process \cite{sethuraman1994constructive}. The generative process of the model (figure \ref{fig:plate})  is given by:
\begin{eqnarray*}
 \beta_k & \sim & \text{Beta}(\cdot ; 1, \eta) \\
                \pi_k & = & \beta_k\prod_{l=1}^{k-1}(1 - \beta_l) \\
 \mathbf{z}_n | \pi& \sim & \textrm{Cat} (\cdot | \pi)  \\
 \bm{\epsilon}_n^{(l)} &\sim& \mathcal{N} (\cdot; \mathbf{0}, \textbf{I} ) \quad \forall l \\
 \mathbf{h}_n^{(L)} &=& m_{\mathbf{z}_n}^{(L)} + s_{\mathbf{z}_n}^{(L)} \odot \bm{\epsilon}_n^{(L)} \\
 \mathbf{h}_n^{(l)} &=& f_{W^{(l)}_{\mathbf{z}_n}}\left(\mathbf{h}_n^{(l+1)}\right) + s_{\mathbf{z}_n}^{(l)} \odot \bm{\epsilon}_n^{(l)} \\
 \mathbf{x}_n | \mathbf{h}_n^{(1)}, \textbf{z}_n &\sim& p_{X} \left(\text{ } \bm{\cdot} \text{ }| f_{W^{(0)}_{\mathbf{z}_n}} \left(\mathbf{h}_n^{(1)}\right) \right)
\end{eqnarray*}
where $\mathbf{h}_n^{(l)} \in \mathbb{R}^{p_l} $ is the $l^{\text{th}}$ layer hidden representation constructed using a nonlinear transformation  $f_{W^{(l)}_{\mathbf{z}_n}}$  represented by a neural network for the cluster assignment $\mathbf{z}_n$. For simplicity, we consider diagonal covariance matrices for each layer where the diagonal elements are $\left[ (s_{\mathbf{z}_n, j}^{(l)}) ^ 2 \right]_{1 \leq j \leq p_l}$,  hence $\odot$ represents the element-wise product. The generalization to full covariance matrices is straightforward using the Cholesky decomposition. 

We denote by $\eta$ the concentration parameter of the Dirichlet process which is a hyperparameter to be tuned manually. The term $p_X$ represents the emission distribution of the observable $\mathbf{x}_n$, usually chosen to be a normal distribution for continuous variables or the Bernoulli distribution for binary variables. We denote the parameters of the generative model by: $$\Theta = \{ m^{(L)}_{1:\infty}, s^{(L)}_{1:\infty}, W^{(0:L-1)}_{1:\infty}, s^{(1:L-1)}_{1:\infty} \} $$
The model thus has an infinite number of parameters due to the Dirichlet process prior. Furthermore, the posterior distribution of the hidden variables cannot be computed in closed-form. In order to perform inference on the model we need to use approximate methods such as Markov Chain Monte Carlo (MCMC) or Variational Inference. MCMC methods are not suitable for high dimensional models such as the DP-DLGMM, where convergence of the Markov chain to the true posterior can prove to be slow and hard to diagnose \cite{blei2017variational}.

In the next section, we develop a structured variational inference algorithm for DP-DLGMM. We show that by choosing a suitable structure for the variational posterior, closed-form solutions can be obtained for the updates of the truncated variational posteriors of the beta distributions, the variational posteriors of the cluster hidden variables, and the optimal prior parameters $\{ m^{(L)}, s^{(L)}\}$ maximizing the evidence lower bound. 

\section{Structured Variational Inference}\label{VI}
For a brief review of variational methods, we denote by $\mathbf{x}_{1:N}$ the $N$ samples present in the dataset supposed to be independent and identically distributed. The log-likelihood of the model is intractable due to the required marginalization of all the hidden variables. In order to bypass this marginalization, we introduce an approximate distribution $q_\Phi$ and use Jensen's inequality to obtain a lower bound \cite{jordan1999introduction}:
\begin{flalign}\label{eq:L}
 & l(\Theta) = \ln p_\Theta ( \mathbf{x}_{1:N} ) \nonumber\\
 & = \ln \left[ \sum_{\mathbf{z}_{1:N}} \int p_\Theta ( \mathbf{x}_{1:N} ,\mathbf{z}_{1:N}, \mathbf{h}^{(1:L)}_{1:N}, \beta )  d \mathbf{h}^{(1:L)}_{1:N} d \beta \right] \nonumber \\
  & \geq  \mathbb{E}_{\mathbf{z}_{1:N}, \mathbf{h}^{(1:L)}_{1:N}, \beta \sim q_\Phi } \left[ \ln \frac{p_\Theta ( \mathbf{x}_{1:N} ,\mathbf{z}_{1:N}, \mathbf{h}^{(1:L)}_{1:N}, \beta )}{q_\Phi(\mathbf{z}_{1:N}, \mathbf{h}^{(1:L)}_{1:N}, \beta | \mathbf{x}_{1:N} )}  \right] \nonumber \\ 
  & \triangleq \mathcal{L}(\Theta, \Phi).
\end{flalign}
We can show that if the distribution  $q_\Phi$ is a good approximation of the true posterior, maximizing the evidence lower bound (ELBO) with respect to the model parameters $\Theta$ is equivalent to maximizing the log-likelihood. For deep generative models, most state-of-the-art methods use inference networks to construct the posterior distribution  \cite{rezende2014stochastic,nalisnick2016stick}. For deep mixture models with discrete latent variables, this approach leads to a mixture density variational posterior where the reparameterization trick requires additional investigation \cite{graves2016stochastic}. Our approach combines standard variational Bayes and neural variational inference. We approximate the true posterior using the following structured variational posterior:
\begin{multline}\label{eq:post}
    q_\Phi (\mathbf{z}_{1:N}, \mathbf{h}^{(1:L)}_{1:N}, \beta | \mathbf{x}_{1:N}) = \prod_{n=1}^N \prod_{l=1}^L q_{\psi^{(l)}_{\textbf{z}_n}} (\textbf{h}^{(l)}_n | \textbf{x}_n, \textbf{z}_n) \\ \times q_{\phi_n} (\textbf{z}_n | \textbf{x}_n)  \prod_{t=1}^T q_{\gamma_t} (\beta_t | \textbf{x}_{1:N}),
\end{multline}
where $T$ is a truncation level for the variational posterior of the beta distributions obtained by supposing that $q(\beta_T = 1) = 1$ \cite{blei2006variational}. We assume a factorized posterior over the hidden layers $\textbf{h}_n^{(1:L)}$, where the intra-layer dependencies are conserved.
\subsection{Deriving the variational posteriors $ q_{\phi_n} $ and $q_{\gamma_t}$}
Deriving the nature of the posterior distributions of the hidden layers $\textbf{h}_n^{(1:L)}$ using the variational approach is intractable due to the nonlinearities present in the model. Thus, we take a similar approach to \cite{rezende2014stochastic}, and we assume that the variational posterior is specified by an inference network, where the parameters of the distribution are the outputs of deep neural networks $\mu_{\psi_t^{(l)}}$ and $\Sigma_{\psi_t^{(l)}}$ of parameters $\psi_t^{(l)}$ for the $l^{\text{th}}$ layer and the $t^{\text{th}}$ cluster:

$$q_{\psi^{(l)}_t} (\textbf{h}^{(l)}_n | \textbf{x}_n, \textbf{z}_n = t) = \mathcal{N} \left( \textbf{h}^{(l)}_n; \mu_{\psi_t^{(l)}} (\textbf{x}_n), \Sigma_{\psi_t^{(l)}} (\textbf{x}_n)\right). $$
In contrast to the hidden layers, we can use the proposed variational posterior of equation \eqref{eq:post} to derive closed-form solutions for $ q_{\phi_n} $ and $q_{\gamma_t}$. Let us consider the  Kullback–Leibler definition of the ELBO $\mathcal{L}$: 
$$ \mathcal{L}(\Theta, \Phi) = - \mathbb{D}_{KL} \left[q_\Phi(\cdot | \textbf{x}_{1:N}) || p_\Theta(\cdot, \textbf{x}_{1:N}) \right]. $$
By plugging the variational posterior and isolating $\beta_t$ terms and $\textbf{z}_n$ terms, we can analytically derive the optimal distributions $q_{\gamma_t}$ and $q_{\phi_n}$ maximizing $\mathcal{L}$:
\begin{eqnarray*}
    q_{\gamma_t} (\beta_t | \textbf{x}_{1:N}) &=& \text{Beta}(\beta_t; \gamma_{1,t}, \gamma_{2,t}) \\
       q_{\phi_n} (\textbf{z}_n | \textbf{x}_{n}) &=& \text{Cat}(\textbf{z}_n; \phi_n),
\end{eqnarray*}
where the fixed point equations for the variational parameters $\phi_n$ and $\gamma_t$ are: 
\begin{flalign}
&\gamma_{1,t} = 1 + \sum_{n=1}^N \phi_{n, t} \label{eq:gamma1} \\
&\gamma_{2,t} = \eta + \sum_{n=1}^N \sum_{r=t+1}^T  \phi_{n, r}  \label{eq:gamma2}
\end{flalign}

\begin{align}
\ln \phi_{n,t} &= \text{const} + \mathbb{E}_{\beta \sim q} [\ln \pi_t ] \nonumber\\ &+ \mathbb{E}_{\textbf{h}_n^{(1:L)} \sim q_{\psi^{(1:L)}_t}} \left[ \ln p_X ( \textbf{x}_n , \textbf{h}^{(1:L)}_n | \textbf{z}_n=t) \right] \nonumber\\ &
+ \sum_l \mathbb{H} \left[ q_{\psi^{(l)}_t}(\cdot | \textbf{z}_n=t, \textbf{x}_n) \right]  \nonumber\\ & \quad \text{s.t.} \quad \sum_{t=1}^T \phi_{n,t} = 1,  \label{eq:phi}
\end{align}
  
The fixed point equation of $\phi_{n,t}$, requires the evaluation of the expectation over the hidden layers, this can be performed by sampling from the variational posterior of each hidden layer and then forwarding the sample using the generative model:
\begin{multline}\label{eq:MC}
    \mathbb{E}_{\textbf{h}_n^{(1:L)} \sim q_{\psi^{(1:L)}_t}} \left[  \ln p_X ( \textbf{x}_n , \textbf{h}^{(1:L)}_n | \textbf{z}_n=t) \right] \\ \approx \frac{1}{S} \sum_{s=1}^S  \ln p_X \left( \textbf{x}_n,  \textbf{h}_{n,t}^{(1:L)(s)} | \textbf{z}_n=t \right) \\
    \text{where: } \quad \textbf{h}_{n,t}^{(l)(s)} \sim q_{\psi^{(l)}_{t}} (\textbf{h}^{(l)}_n | \textbf{x}_n, \textbf{z}_n = t).
\end{multline}

A key insight here is the following: if a cluster $t$ is incapable of reconstructing a sample $\textbf{x}_n$ from the variational posterior, this will reinforce the belief that $\textbf{x}_n$ should not be assigned to that cluster. Furthermore, the estimation of the expectation can be performed using the same reparameterization trick that we will develop in section \ref{reparam}.

\subsection{Closed-Form  updates for $m^{(L)}_{1:T}$ and $s^{(L)}_{1:T}$ }
In addition to the variational posteriors of the beta distributions and the cluster assignments, closed-form solutions can be obtained for the updates of $m^{(L)}_{1:T}$ and $s^{(L)}_{1:T}$. Let us reconsider the evidence lower bound of equation \eqref{eq:L}, where we isolate only terms dependent on the prior parameters. We have:
\begin{multline*}
    \mathcal{L}(m^{(L)}_{1:T}, s^{(L)}_{1:T}) = \text{const} -  \sum_{n,t} \phi_{n,t} \\ \times \mathbb{D}_{KL} \left[ \mathcal{N}(\mu_{\psi_t^{(L)}}( \textbf{x}_n), \Sigma_{\psi_t^{(L)}}( \textbf{x}_n) ) || \mathcal{N}( m^{(L)}_{t}, V^{(L)}_t ) \right].
\end{multline*}
where $V^{(L)}_t = \text{diag}\left[ (s_{t, j}^{(L)}) ^ 2 \right]_{1 \leq j \leq p_L} $ represents the covariance matrix of the $L^{\text{th}}$ layer. By setting the derivative of $\mathcal{L}$ with respect to the parameters to zero, we obtain:
\begin{equation}\label{eq:m}
    m_t^{(L)} = \frac{1}{N_t} \sum_{n=1}^N \phi_{n,t} \mu_{\psi_t^{(L)}}( \textbf{x}_n) \quad N_t = \sum_{n=1}^N \phi_{n,t}
\end{equation}
\begin{multline}\label{eq:V}
    V_t^{(L)} = \frac{1}{N_t}  \sum_{n=1}^N \phi_{n,t} \textbf{I} \odot \Big\{ \Sigma_{\psi_t^{(L)}}( \textbf{x}_n) \\ + (\mu_{\psi_t^{(L)}} (\textbf{x}_n) - m_t^{(L)}) (\mu_{\psi_t^{(L)}} (\textbf{x}_n) - m_t^{(L)})^{\text{T}} \Big\},
\end{multline}
where to extract the diagonal elements we perform an elementwise multiplication by the identity matrix $\textbf{I}$. The update rules obtained are similar to the M-Step of a classical Gaussian Mixture Model, except in this case the updates are performed on the last hidden layer of the generative model, and the E-step of equation \eqref{eq:phi} takes into account all the hidden layers. Detailed derivation of the previous equations are presented in the supplementary material.
\subsection{Stochastic Backpropagation}\label{reparam}
We next show how to perform stochastic backpropagation in order to maximize $\mathcal{L}$ with respect to the parameters $\psi$ and $\Lambda = \{W^{(0:L-1)}_{1:T}, s^{(1:L)}_{1:T} \}$. Similarly to the previous section, we isolate the terms in the evidence lower bound dependent on $\psi$ and $\Lambda$. We have:

\begin{multline}\label{eq:LW}
    \mathcal{L}(\psi, \Lambda) = \text{const} +  \sum_{n,t} \phi_{n,t} \Big\{ \sum_l \mathbb{H} \left[ q_{\psi^{(l)}_t}(\cdot | \textbf{z}_n=t, \textbf{x}_n) \right] \\ + \mathbb{E}_{\textbf{h}_n^{(1:L)} \sim q_{\psi^{(1:L)}_t}} \left[  \ln p_X ( \textbf{x}_n , \textbf{h}^{(1:L)}_n | \textbf{z}_n=t) \right]  \Big\}.
\end{multline}

By taking the expectation over the hidden cluster variables $\textbf{z}_n$, we obtain conditional expectations over the hidden layers $\textbf{h}^{(1:L)}_n$ knowing the cluster assignment. In order to backpropagate gradients of $\Lambda$ and $\psi$, it suffices to perform a reparameterization trick for each cluster assignment at each hidden layer (proof in Appendix \ref{reparame}). We can achieve this by sampling:
$$ \epsilon_{n, t}^{(l)} \sim \mathcal{N}(\textbf{0},\textbf{I}), $$
a sample from the posterior of the $l^{\text{th}}$ hidden layer can then be obtained by the following transformation:
$$ \textbf{h}_{n, t}^{(l)} = \mu_{\psi_t^{(l)}} (\textbf{x}_n) + \epsilon_{n, t}^{(l)} \sqrt{\Sigma_{\psi_t^{(l)}}( \textbf{x}_n)}, $$
where $\Sigma_{\psi_t^{(l)}}( \textbf{x}_n)$ is supposed to be a diagonal matrix for simplicity. Following the previous analysis, we can derive an algorithm to perform inference on the proposed model, where between iterations of the fixed point update steps, $E$ epochs of gradient ascent are performed to obtain a local maximum of the ELBO with respect to $\Lambda$ and $\psi$. Algorithm \ref{alg:the_alg} summarizes the process.
\begin{algorithm}[t]
\caption{Variational Inference for the DP-DLGMM}
\begin{algorithmic}
\STATE \text{\textbf{Input:}} $\textbf{x}_{1:N}, T,  \eta, \alpha$ 
\STATE \text{Initialize} $\phi, \Lambda, \psi$ 
\WHILE{\text{not converged}}
\STATE $\text{update: } \gamma_{t} \quad \forall t $ \COMMENT{\eqref{eq:gamma1}, \eqref{eq:gamma2}}
\STATE $\text{update: } m^{(L)}_{t} \quad \forall t $ \COMMENT{\eqref{eq:m}}
\STATE $\text{update: } V^{(L)}_{t} \quad \forall t $ \COMMENT{\eqref{eq:V}}
\FOR{\text{each epoch}}
\STATE $ \Lambda \leftarrow \Lambda +  \alpha \partial_\Lambda \mathcal{L}$
\STATE $ \psi \leftarrow  \psi + \alpha \partial_ \psi \mathcal{L}$
\ENDFOR
\STATE $\text{update: } \phi_{n,t}  \quad \forall n, \forall t$ \COMMENT{\eqref{eq:phi}}
\ENDWHILE
\end{algorithmic}
\label{alg:the_alg}
\end{algorithm}

\section{Semi-Supervised Learning (SSL)}\label{SSL}

\subsection{SSL using the DP-DGLMM}
In this section, similarly to \cite{kingma2014semi} we consider a partially labeled dataset $\textbf{x}_{1:N} = D_l \cup D_u$, where $D_l = \{ \textbf{x}_n, \textbf{y}_n \}_n $ is the labeled part, $\textbf{y}_n$ represents the label of the sample $\textbf{x}_n$, and $D_u$ represents the unlabeled part. The log likelihood can be divided for the labeled and unlabeled parts as:
\begin{eqnarray*}
    l(\Theta) & = & \ln p_\Theta ( \mathbf{x}_{1:N} ) \nonumber\\
    & = & \sum_{\mathbf{x}_n \in  D_l } \ln p_\Theta (\mathbf{x_n}) + \sum_{\mathbf{x}_n \in  D_u } \ln p_\Theta (\mathbf{x}_n) \nonumber \\
    & = & \sum_{\mathbf{x}_n \in  D_l } \ln p_\Theta (\textbf{x}_n, \textbf{z}_n = \textbf{y}_n ) + \sum_{\mathbf{x}_n \in  D_u } \ln p_\Theta (\mathbf{x}_n). \\
\end{eqnarray*}
The last equation follows from the fact that $ p_\Theta (\textbf{x}_n | \textbf{z}_n \neq \textbf{y}_n ) = 0$. By dividing the labeled and unlabeled parts of the dataset, we can follow the same approach presented in section \ref{VI} in order to derive a variational inference algorithm. In this case, the fixed point updates and the gradient ascent steps remain unchanged if we set $\phi_{n,\textbf{y}_n} = 1$ for a labeled $\textbf{x}_n$ sample.

\subsection{The predictive distribution}
In order to make predictions using the model, we need to evaluate the predictive distribution. Given a new sample $\textbf{x}_{N+1}$, the objective is to evaluate the following quantity $p(\textbf{z}_{N+1} = k | \textbf{x}_{1:N+1})$. This task requires an intractable marginalization over all the other hidden variables. However, similarly to \cite{blei2006variational}, we can use the variational posterior to approximate the true posterior, which in turn leads to simpler expectation terms:
\begin{flalign}
p(\textbf{z}_{N+1} & = k  | \textbf{x}_{1:N+1}) \propto p(\textbf{z}_{N+1} = k , \textbf{x}_{N+1} | \textbf{x}_{1:N}) \nonumber\\
& \approxprop \mathbb{E}_{\beta \sim q} \left [ \pi_k(\beta) \right]\nonumber \\  \times & \mathbb{E}_{\textbf{h}^{(1:L)}_{N+1} \sim q_{\psi_k^{(1:L)}}} \left[ p_X \left( \textbf{x}_{N+1} | f_{\Lambda} ( \textbf{h}^{(1:L)}_{N+1}), \textbf{z}_n=k  \right) \right] \label{eq:pred}
\end{flalign}
where  $f_{\Lambda} (\cdot)$ represents the forward pass over the generative model. The expectation with respect to the beta terms can be computed in closed-form as a product of expectations over the beta posteriors. The second expectation can be evaluated using the Monte-Carlo estimator of equation \eqref{eq:MC}.

\section{Experiments}
\subsection{Evaluation of the semi-supervised classification}

\begin{figure}[t]
    \centering
    \includegraphics[width=0.48\textwidth]{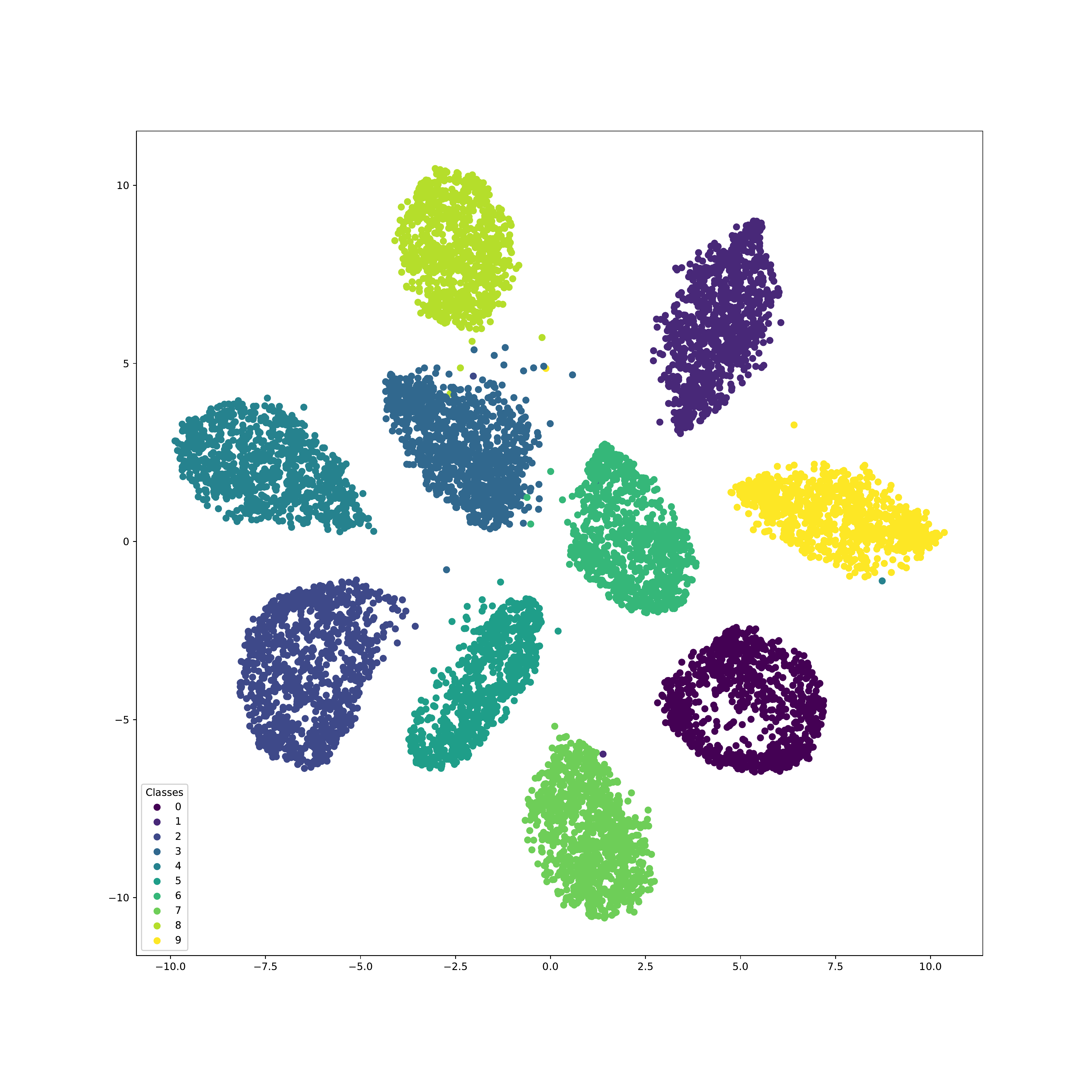}
    \caption{t-SNE plot of the second stochastic hidden layer on the MNIST test set for the semi-supervised (10\% labels) version of the DP-DLGMM.}
     \label{fig:tsnessl}
\end{figure}

We evaluate the semi-supervised classification capabilities of the model. We train our DP-DLGMM model on the MNIST dataset \cite{lecun-mnisthandwrittendigit-2010} with train-valid-test splits equal to $\{45000, 5000, 10000\}$ similarly to \cite{nalisnick2016stick}, with 10 \% labelisation randomly drawn. We run the process for 5 iterations, and we evaluate our model on the test set. We report the mean and standard deviation of the classification error in percentages in Table \ref{table:class}. Our method produces a  competitive score with existing state-of-the art methods: Deep Generative Models (DGM) \cite{kingma2014semi} and Stick-Breaking Deep Generative Models (SB-DGM) \cite{nalisnick2016stick}. Unlike the previous approaches, the loss was not up-weighted for the labeled samples. Figure \ref{fig:tsnessl} shows the t-SNE projections \cite{maaten2008visualizing} obtained with 10 \% of the labels provided. We notice that by introducing a small fraction of labels the class structure was highly preserved in the latent space.

\begin{figure}[t]
    
    \centering
    \includegraphics[width=0.48\textwidth]{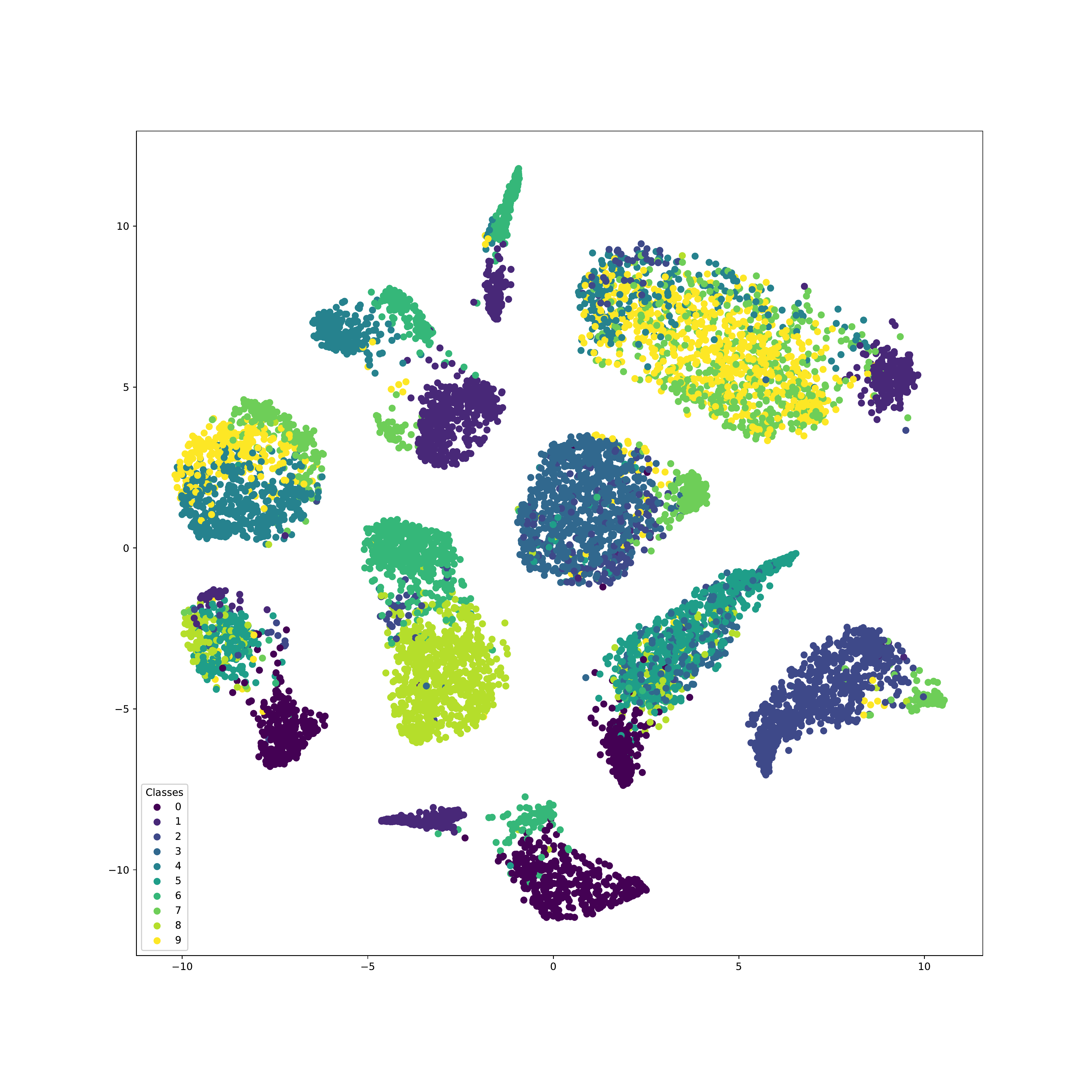}
    \caption{t-SNE plot of the second stochastic hidden layer on the MNIST test set for the unsupervised version of the DP-DLGMM.}
    \label{fig:tsneunsup}
\end{figure}

\begin{table}[h!]
  \centering
  \resizebox{\columnwidth}{!}{\begin{tabular}{llll}
    \toprule 
    \makecell{kNN \\ (k=5) } & DGM & SB-DGM  & DP-DLGMM \\ 
    \midrule
      $6.13 \pm .13$ &    $4.86 \pm .14$ &  $3.95 \pm .15  $   &  $   \mathbf{2.90 \pm .17} $ \\
   \bottomrule
 \end{tabular}}
 \caption{Semi-supervised classification error (\%) on the MNIST test set with 10 \% labelisation. Comparison with \cite{nalisnick2016stick}}. \label{table:class}
\end{table}

\subsection{Data generation and visualization}

\begin{figure*}
    \centering
    \includegraphics[width=0.6\textwidth]{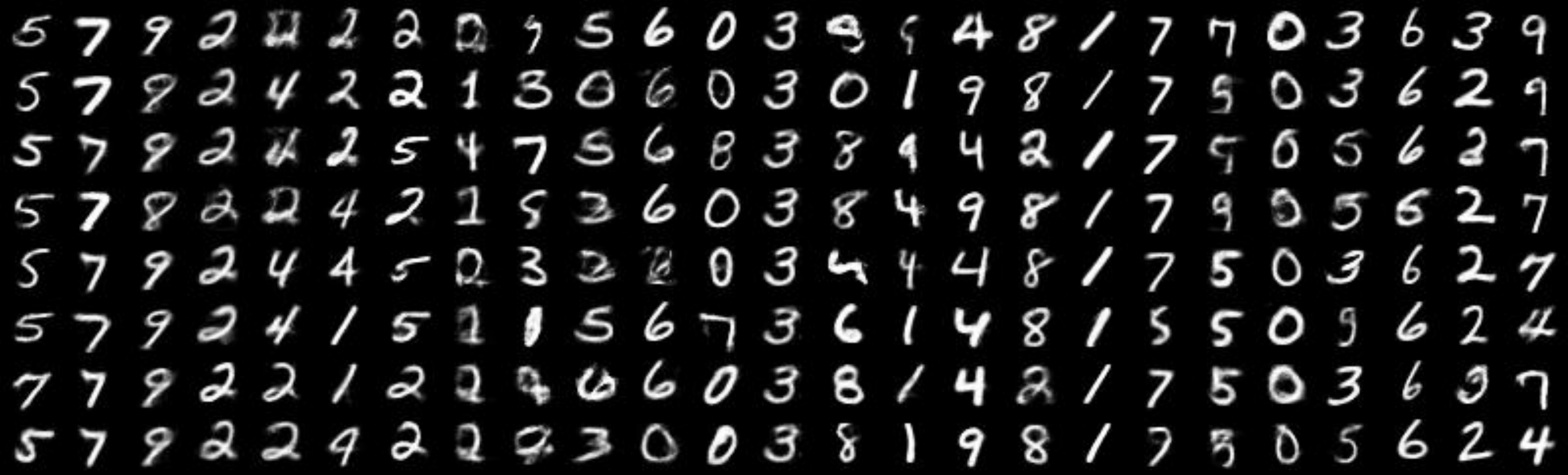}
    \includegraphics[width=0.25\textwidth]{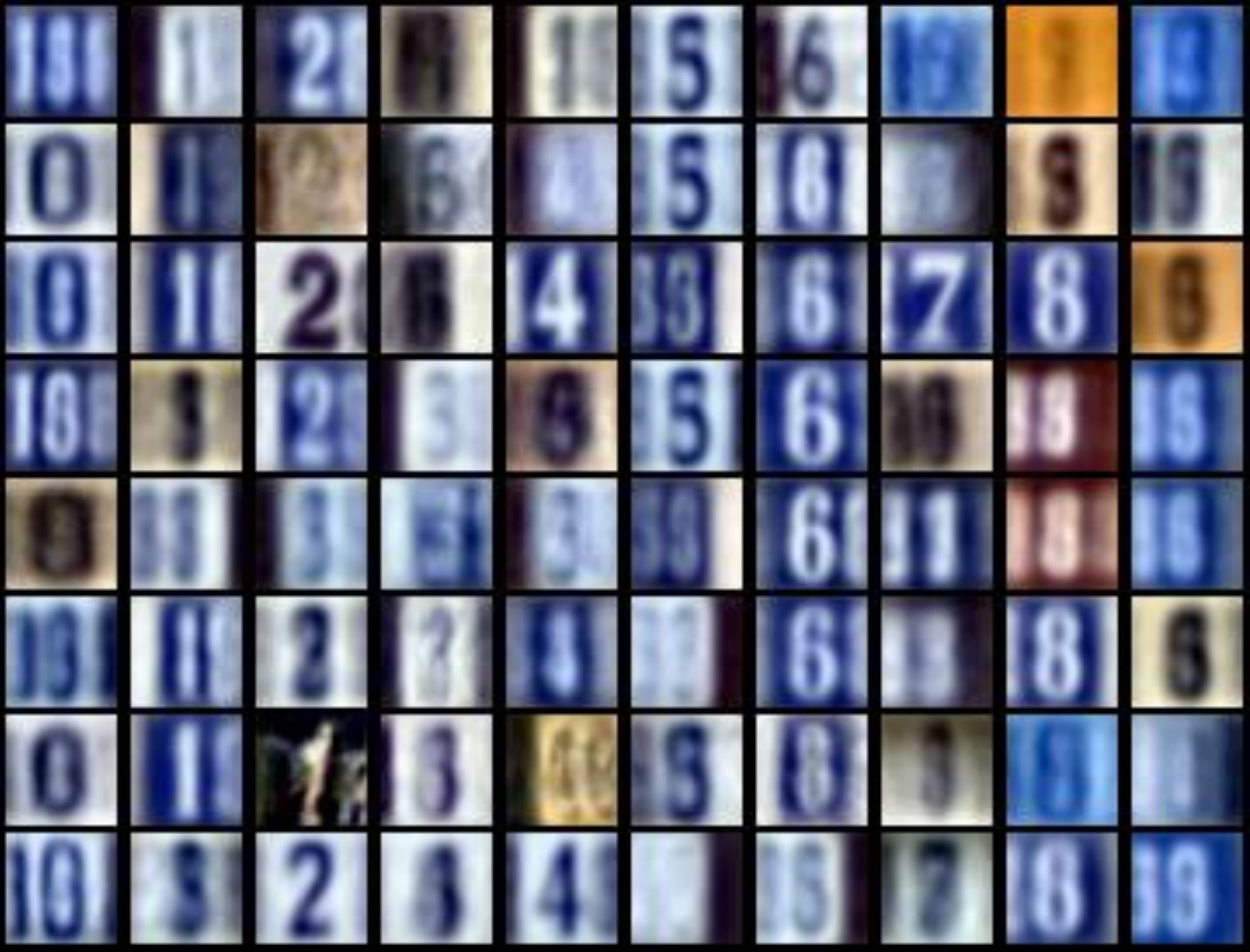}
    \caption{Generated samples from the DP-DLGMM model for the unsupervised version on the MNIST dataset (left) and the semi-supervised version on the SVHN dataset (right).}
    \label{fig:samples}
\end{figure*}

To further test our model, we generate samples for each cluster from the models trained on both the MNIST and SVHN \cite{netzer2011reading} datasets. The MNIST model is trained in an unsupervised manner, and the SVHN model is trained with semi-supervision where we provide 1000 randomly generated labels. The samples obtained are represented in figure \ref{fig:samples}. For the unsupervised model, we notice that the clusters are representative of the shape of each digit. We plot the t-SNE projections of the MNIST test set of the unsupervised model in Figure \ref{fig:tsneunsup}. We notice that the digits belonging to the same true class tend to group with each other. However, two groups of the same class can be very separated in the embedding space. The interpretation we can draw from this effect is that the DP-DLGMM tends to separate the latent space in order to distinguish between the variations of hidden representations of the same class. The clusters obtained are not always representative of the true classes which is a common effect with infinite mixture models. In a full unsupervised setting, data can be explained by multiple correct clusterings. This effect can simply be countered by  adding a small supervision (figure \ref{fig:tsnessl}).

\section{Conclusion}

In this paper, we have presented a variational inference method for Dirichlet Process Deep Latent Gaussian Mixture Models. Our approach combines classical variational inference and neural variational inference. The algorithm derived is thus a standard variational inference algorithm, with fixed point updates over a subset of the parameters presenting linear dependencies. The parameters present in nonlinear transformations are updated using standard gradient ascent where the reparameterization trick can be applied for the variational posterior of the stochastic hidden layers knowing the cluster assignments. Our approach shows promising results both for the unsupervised and semi-supervised cases. In future work, stochastic variational inference can be explored to speed-up the training procedure. Our approach can also be generalized to other types of deep probabilistic graphical models.

\appendix

\section{Proof of the reparameterization trick knowing the cluster assignment}\label{reparame}
The evidence lower bound of our model can be written in its general form as:
\begin{eqnarray*}
\mathcal{L}(\theta, \psi) &=& \sum_{t=1}^T \phi_t \times \mathbb{E}_{h \sim \mathcal{N}(\mu_{\psi_t}(x), \sigma_{\psi_t}(x)^2 \textbf{I}) } \left[ f_\theta (h) \right] \\
&=& \sum_{t=1}^T \phi_t \int_h \mathcal{N}(h, \mu_{\psi_t}(x), \sigma_{\psi_t}(x)^2 \textbf{I}) f_\theta (h) dh \\
&=& \sum_{t=1}^T \phi_t \int_h \mathcal{N}(h, \mu_{\psi_t}(x), \sigma_{\psi_t}(x)^2 \textbf{I}) \\ & \times &f_\theta (h) | \partial_{\epsilon} h_t | d\epsilon.
\end{eqnarray*}
By introducing the following transformation: $$ h_t(\epsilon) = \mu_{\psi_t}(x) + \sigma_{\psi_t}(x) \odot \epsilon \quad \epsilon \sim \mathcal{N}(\textbf{0},\textbf{I}),$$
and using the density transformation lemma:
$$  \mathcal{N}(h, \mu_{\psi_t}(x), \sigma_{\psi_t}(x)^2 \textbf{I})  | \partial_{\epsilon} h_t | =  \mathcal{N}(\epsilon; \textbf{0},\textbf{I}), $$
 we have:
\begin{eqnarray*}
\mathcal{L}(\theta, \psi)& = & \sum_{t=1}^T \phi_t  \mathbb{E}_{\epsilon} \left[  f_\theta (\mu_{\psi_t}(x) + \sigma_{\psi_t}(x) \odot \epsilon) \right] \\ 
&\approx&  \sum_{t=1}^T \phi_t f_\theta (\mu_{\psi_t}(x) + \sigma_{\psi_t}(x) \odot \hat{\epsilon}).
\end{eqnarray*}

where $\hat{\epsilon}$ is a sample drawn from $\mathcal{N}(\textbf{0},\textbf{I})$, thus we can backpropagate stochastic gradients for each class assignment. 

\section{Stochastic Variational Inference}\label{SVB}
Updating the $\Lambda$ and $\psi$ parameters using $E$ epochs of gradient ascent significantly adds to the complexity of Algorithm \ref{alg:the_alg}. One possible approach is to perform stochastic variational inference \cite{hoffman2013stochastic} for fixed point update equations. This allows for the use of the same batch of data for the gradient ascent steps of $\Lambda$ and $\psi$ and the stochastic updates of the fixed point equations. Let us consider a batch $\textbf{x}_{1:B}$, the updates in this case are:
\begin{eqnarray*}
\gamma^{(t+1)}_{1,k} &=& (1 - \rho_t) \gamma^{(t)}_{1,k} + \rho_t \frac{N}{B} \hat{\gamma}_{1,k} \\
\gamma^{(t+1)}_{2,k} &=& (1 - \rho_t) \gamma^{(t)}_{2,k} + \rho_t \frac{N}{B} \hat{\gamma}_{2,k} \\
m^{(t+1)}_{k} &=& (1 - \rho_t) m^{(t)}_{k} + \rho_t \frac{N}{B} \hat{m}_{k} \\
V^{(t+1)}_{k} &=& (1 - \rho_t) V^{(t)}_{k} + \rho_t \frac{N}{B} \hat{V}_{k},
\end{eqnarray*}
where $\hat{\gamma}_{1,k}$, $\hat{\gamma}_{2,k}$, $\hat{m}_{k}$, $\hat{V}_{k}$, are computed for the minibatch $\textbf{x}_{1:B}$  using equations \eqref{eq:gamma1},\eqref{eq:gamma2},\eqref{eq:m}, and \eqref{eq:V} respectively. In order to guarantee convergence $\rho_t$ must satisfy: 
$$\sum_t  \rho_t = \infty \quad \text{and} \quad \sum_t  \rho_t ^2 < \infty. $$
\bibliography{example_paper}
\bibliographystyle{icml2019}

\end{document}